\documentclass{article}
\usepackage[utf8x]{inputenc}
\usepackage{listings}
\usepackage{graphicx}
\usepackage{mathtools}
\usepackage{amsmath}
\usepackage{amssymb}
\usepackage{placeins}
\usepackage{subfig}
\usepackage{imakeidx}
\usepackage{hyperref}

\makeindex[intoc]

\newcommand{\symbolnomenclature}[2]{\indent\parbox{30mm}{#1}{#2\par}}

\begin{document}

\title{GRD-Net: Generative-Reconstructive-Discriminative Anomaly Detection with Region of Interest Attention Module}

\date{06/12/2022}

\author{Niccolò Ferrari\thanks{Department of Engineering, University of Ferrara, Via Saragat 1, 44122 Ferrara, Italy} \thanks{Bonfiglioli Engineering, Via Amerigo Vespucci 20, 44124 Ferrara, Italy} \thanks{niccolo.ferrari@unife.it, nferrari@bonfiglioliengineering.com}\and Michele Fraccaroli\footnotemark[1]\and Evelina Lamma\footnotemark[1]}

\maketitle

\section{Abstract}
	\label{sec: abstract}
	{
		Anomaly detection is nowadays increasingly used in industrial applications and processes. One of the main fields of the appliance is the \textit{visual inspection} for surface anomaly detection, which aims to spot regions that deviate from regularity and consequently identify abnormal products.
		Defect localization is a key task, that usually is achieved using a basic comparison between generated image and the original one, implementing some blob-analysis or image-editing algorithms, in the post-processing step, which is very biased towards the source dataset, and they are unable to generalize. Furthermore, in industrial applications, the totality of the image is not always interesting but could be one or some regions of interest (ROIs), where only in those areas there are relevant anomalies to be spotted.
		For these reasons, we propose a new architecture composed by two blocks.
		The first block is a Generative Adversarial Network (GAN), based on a residual autoencoder (ResAE), to perform reconstruction and denoising processes, while the second block produces image segmentation, spotting defects. This method learns from a dataset composed of good products and generated synthetic defects.
		The discriminative network is trained using a ROI for each image contained in the training dataset. The network will learn in which area anomalies are relevant. This approach guarantees the reduction of using pre-processing algorithms, formerly developed with blob-analysis and image-editing procedures.
		To test our model we used challenging MVTec anomaly detection datasets and an industrial large dataset of pharmaceutical BFS strips of vials. This set constitutes a more realistic use case of the aforementioned network.	
	}

\section{Keywords}
	{
		Anomaly Detection, Attention Module, Generative Adversarial Network, Defect Localization, Region of Interest
	}
	
\section{Acknowledgments}
	\label{sec: acknowledgments}
	{
		The authors would like to thank Bonfiglioli Engineering for providing a real-case dataset to test the software developed in this work. The first author is supported by a industrial PhD funded by Bonfiglioli Engineering, Ferrara, Italy.
		The other author is supported by a PhD scholarship funded by the Emilia Romagna region, Italy, under POR FSE 2014–2020 program.
	}

{
	\section{Nomenclature}
		\label{sec: nomenclature}
		{
			\noindent
			
			\symbolnomenclature{AE}{AutoEncoder}
			\symbolnomenclature{VAE}{Variational AutoEncoder}
			\symbolnomenclature{CNN}{Convolutional Neural Network}
			\symbolnomenclature{RNN}{Recurrent Neural Network}
			\symbolnomenclature{LSTM}{Long Short Time Memory}
			\symbolnomenclature{GAN}{Generative Adversarial Network}
			\symbolnomenclature{Generator}{Generative subnet of the GAN}
			\symbolnomenclature{Discriminator}{Adversarial subnet of the GAN}
			\symbolnomenclature{Discriminative net}{U-Net subsequent to the GAN used for segmentation}
			\symbolnomenclature{CRAE}{fully-Convolutional Residual AutoEncoder}
			\symbolnomenclature{DRAE}{Dense-bottleneck Residual AutoEncoder}
			\symbolnomenclature{AUROC}{Area Under the Receiver Operating Characteristic}
			\symbolnomenclature{ROI}{Region Of Interest}
			\symbolnomenclature{SSIM}{Structural Similarity Index Measure}
		}
}

\section{Introduction}
	\label{sec: intro}
	{
		Semi-supervised computer vision is a task increasingly used in the industrial sector. The reasons are to be found in its flexibility and in its capability to generalize when a new anomaly is seen. Moreover, despite the good performance of supervised approaches in computer vision fields, but requiring a large number of examples during the training phase, the previously mentioned approach requires only a significant number of nominal examples to define their distribution. Regarding defects, on the other hand, it requires just a little set of anomalies, used for testing purposes and in some cases to define an anomaly threshold.
		
		\medskip
		In real cases, on production lines, the availability of regular products is the vast majority compared to anomalies. For this reason, the training dataset would be extremely unbalanced in favor of nominal examples. This makes it difficult performing a good training supervised model. Moreover is often required to locate the defect within the image, because usually, the defective portion covers a small area of the whole surface. To give an example, on pharmaceutical vials defect are frequently little scratches, very small black spots or some alien particles deposited on the surface of products, which are usually between 100 and 1000 ${\mu}m$. This target could be more easily reached with a semi-supervised anomaly detection architecture, and more specifically with a reconstruction-based approach, in which the network gives you a reconstruction of the image without the anomalous area.
		
		\medskip
		Reconstructive methods include Autoencoder (AE) \cite{DBLP:journals/corr/abs-2003-05991,DBLP:journals/corr/abs-2201-03898}, Variational Autoencoder (VAE) \cite{DBLP:journals/corr/abs-1906-02691,https://doi.org/10.48550/arxiv.1312.6114} and Generative Adversarial Network (GAN) \cite{goodfellow2020generative}.
		They have been thoroughly investigated since they make it possible to learn a robust reconstruction subspace using only images without anomalies.
		Thanks to the incapability to rebuild anomalous regions, which was not contained within nominal images during training, the network fails to reproduce out-of-distribution area.
		For this reason, it's possible to detect discrepancies between the two images by thresholding, for example, the absolute value of the difference between them.
		This is the most immediate and simple method for performing final classification, but it is a non-parametric approach for anomaly localization, and in some cases, discrimination could be erroneous in some noisy cases, because the sum of all the small differences could exceed the threshold. In other cases could be inaccurate, due to the lack of comprehension of differences between the two images, which is left to a simple threshold.
		
		\medskip
		In addition to this, our \textit{Generative-Reconstructive-Discriminative Anomaly Detection with Region of Interest Attention Module} network (\textit{GRD-Net}) aims to minimize the development of pre-processing algorithms that are used to locate the portion of the image in which searching for anomalies.
		In classical \emph{reconstruction-based} methods, the generation of anomaly map, as mentioned above, is left to a threshold-based classifier, so is not possible to turn attention to one or more specific ROIs.
		\emph{Embedding similarity-based} approach constitutes another big family of architectures, offering encouraging results. However, due to the lack of comprehensibility of the result and learning process, it becomes more difficult to impose an ROI, which draws attention to defects.
		
		\medskip
		For all the aforementioned reasons, a second network, chained to the first generative part, is required to achieve the intended results.
		This work is heavily inspired by the discriminatively trained reconstruction anomaly embedding model (DR\AE M) \cite{zavrtanik2021draem}. DR\AE M works by learning a joint representation of an anomalous image and its anomaly-free reconstruction and, simultaneously learning a decision boundary between anomalous and positive examples. This method enables direct anomaly localization avoiding the implementation of some post-processing techniques. 
		
		\medskip
		DR\AE M is based on a first \emph{reconstructive network} (an autoencoder) and a \emph{discriminative network}. The first network is trained to identify and reconstruct anomalies, maintaining the non-anomalous regions of the input image. The second network combines original and reconstructed appearance to learn joint-anomaly inclusion reconstruction, to produce accurate anomaly segmentation maps \cite{zavrtanik2021draem}.
		
		\medskip
		In the context of this work, the autoencoder that defines the reconstructive network is replaced with GANomaly \cite{akcay2018ganomaly}. GANomaly is a Generative Adversarial Network (GAN) \cite{goodfellow2020generative} architecture that simultaneously learns how to create a high-dimensional image space and infer a latent space. The model can map the input image to a lower dimension vector using encoder-decoder-encoder sub-networks, which are then used to reconstruct the generated output image.
		This generated image is mapped to its latent representation by the additional encoder network. Learning the data distribution for the normal samples is aided by minimizing the distance between these images and the latent vectors during training.
		The generative part of the GAN is constituted by a fully-convolutional residual autoencoder \cite{articleWickramasinghe}, that, as mentioned by \emph{Wickramasinghe et al.}, residual blocks help to prevent the gradient vanishing on deep convolutional networks, and thus avoiding the deterioration of learned embedded-representations.
		
		{
			\medskip
			In this work, summarizing:
			\begin{enumerate}
				\item the generalization capability of the GANomaly architecture \cite{akcay2018ganomaly} with the denoising ability derived from the DR{\AE}M architecture \cite{zavrtanik2021draem} are merged in the reconstructive part of the model
				\item the reconstructive autoencoder is residual and fully convolutional \cite{articleWickramasinghe}, improving the stability of the learning process
				\item an \emph{attention module} that uses a ROI is added for each example during the training phase, to learn the area where to focus the segmentation of the abnormal area, in the discriminative part of the model
			\end{enumerate}
		}

		{
			\medskip
			The first network rebuilds the original image in a better and more precise way with a more performing and stable training phase, regarding the two reference models GANomaly and DR{\AE}M. This is thanks to the residual autoencoder with the GAN structure and the mask superimposed, obtained by adding Perlin noise to the input. This technique challenges the network not only to rebuild the input image as it is, but also to regenerate the hidden part by the noise in a coherent way.
			The second block identifies the area where the defect is located, which is a specification required in most industrial applications, with an attention module ROI-based. Defining a ROI for each training example lets the network learn the important area of the product where to look for defects, using the original and the reconstructed image by the first block. In this way, the second net generalizes and spots the ROI in the new input images during production, excluding the research of defects outside. This is a very important result because often we need to spot defects within a region of interest (ROI), excluding the more chaotic and false-reject-prone area outside.
		}
		
		\medskip
		The rest of the paper is organized as follows: Section \ref{sec: related} describes related works. Section \ref{sec: methods} present the background knowledge necessary for the correct understanding of this work (Sections \ref{ch: draem} and \ref{ch: ganomaly}) and our contribution (\ref{ch: draem_gan}). Section \ref{sec: experiments} illustrates the experiments and the results obtained on the various datasets. Finally, in Section \ref{sec: conc_future} we present conclusions and future work.
	}

\section{Related Work}
	\label{sec: related}
	{
		Many surface anomaly detection techniques exploit the \emph{reconstruction-based} approach. This approach is based on image reconstruction and identify anomalies working on image reconstruction error \cite{akcay2018ganomaly, akccay2019skip, bergmann2018improving}. Typically, neural networks like Autoeconders (AEs) \cite{bergmann2018improving, gong2019memorizing}, Variational Autoencoders (VAEs) \cite{venkataramanan2020attention} and Generative Adversarial Networks (GANs) \cite{akcay2018ganomaly, pidhorskyi2018generative, sabokrou2018adversarially, XIA2022497} are used for image reconstruction as described in \cite{bergmann2018improving}. The finding of an anomaly is generally based on the quality of image reconstruction. Reconstruction-based methods can use the structural similarity \cite{bergmann2018improving} or the pixel-wise reconstruction error \cite{bergmann2019mvtec} as the anomaly score to localize anomalies. A visual attention map created from the latent space can also be used as the anomaly map \cite{venkataramanan2020attention}. Another reconstruction-based model that implement a segmentation structure based on transformer is RDAD \cite{https://doi.org/10.1002/int.22974}. The reconstruction-based approach are easily interpretable but their performance is constrained by the fact that AE can occasionally produce good reconstruction outcomes for anomalous images as well \cite{perera2019ocgan}. A good comparison and analysis of the different techniques was described by Xuan Xia et. al. \cite{XIA2022497}, explaining the benefits of a semi-supervised machine learning architecture for \emph{reconstruction-based} method, but also comparing some \emph{embedding similarity-based} methods.
		
		Another important family of methods in the filed of anomaly detection is, precisely, the \emph{embedding similarity-based} approach. These techniques extract useful vectors describing an entire image for anomaly detection \cite{rippel2021modeling, bergman2020deep} or an image patch for anomaly localization \cite{napoletano2018anomaly} using deep neural networks. However, in several works based on embedding similarity-based methods, it offers encouraging results but frequently lacks interpretability. It is impossible to identify the specific aspect of an anomalous image that contributed to its anomaly score. The anomaly score is in this case the distance between the embedding vectors of a test image and the reference vectors representing normality from the training dataset. The normal reference can be defined as the center of a sphere that contains embedding from normal images or the entire set of normal embedding as in the case of SPADE \cite{cohen2020sub}. Another interesting approach that work with patch embedding is PaDiM \cite{defard2021padim}. Normal class in PaDiM is described through a set of Gaussian distributions that utilizes the pre-trained Convolutional Neural Network (CNN) to models correlations between semantic levels. Heavily related to SPADE and PaDiM, there is PatchCore \cite{roth2022towards} that uses a memory bank with neighbourhood-aware patch-level features in order to increase performance. Additionally, corset sub-sampling of the memory bank ensures low inference cost at higher performance. 
		A further sub-category of methods, however based on embedding similarity-based approach, is the one based on generative models called \emph{normalizing flows} (NFLOW) \cite{dinh2016density}. The main advantage of NFLOW models is ability to estimate the exact likelihoods for out-of-distribution examples compared to other generative models \cite{shi2021unsupervised, schlegl2017unsupervised, schlegl2019f}. Notable works in the NFLOW category can be the system developed by Rudolph et. al. called DifferNet \cite{rudolph2021same}, the work of Gudovskiy et. al. called CFLOW-AD \cite{gudovskiy2022cflow} and the more recent Jaehyeok Bae et. al. work called PNI \cite{bae2023pni}, which takes into account the position and neighborhood information on the distribution of normal features.
			
			\emph{Knowledge distillation} techniques are also widely used in anomaly detection tasks, especially when we're dealing with large images, as in the work of Paul Bergmann et. al. \cite{DBLP:journals/corr/abs-1911-02357}. This matter is examined in the work also written by Paul Bergmann et. al. \textit{Beyond Dents and Scratches} \cite{Bergmann2022}, in which anomalies are divided into logical and structural. Noteworthy is also the knowledge distillation-based work of Kilian Batzner et. al. EfficientAD \cite{batzner2023efficientad} where processing time plays a central role in the problem definition, because more and more often lots of real-time applications use unsupervised machine learning algorithms for anomaly detection tasks.
		
		\emph{Reconstruction-based} anomaly detection approaches are widely used in different areas with other types of data, like time series data. In these cases conventional threshold-based anomaly detection methods are inadequate, as mentioned by Dan Li et. al. \cite{li2019madgan}. To handle this type of data, an LSTM-RNN model must be introduced into the GAN or VAE-GAN architecture \cite{Zhuinproceedings, Niuarticle} with an encoder-decoder-encoder shape.
			Such data may be derived from industrial processes \cite{Jiangarticle}, where it is often difficult to obtain balanced data between regular and abnormal data. Also, it can be obtained by smart grids \cite{Radoglouarticle, Siniosoglouarticle}, where it is mandatory to monitor data for security tasks but equally difficult to handle such big data without artificial intelligence algorithms; finally, data could consist of video streams \cite{CHEN2021107969}.
		
		A special mention should be made to the work of Zavrtanik et. al. \cite{zavrtanik2021draem} as this work is largely based and inspired by DR\AE M. This work exploits a reconstruction and a discriminative network to segment artificial noise. The output of DR\AE M is an anomaly detection mask and the anomaly score. The anomaly mask can be used to estimate the image-level anomaly score. The maximum value of the smoothed anomaly score map is used to calculate the final score.
	}

\section{Methods}
	\label{sec: methods}
	{
		To explain our approach, we firstly introduces DR\AE M and GANomaly as they are the knowledge base necessary for understanding the rest of the paper.
		
		{
			\medskip
			In this section, summarizing:
			\begin{enumerate}
				\item DR{\AE}M architecture \cite{zavrtanik2021draem}, which is the starting point of our improvements
				\item GANomaly architecture \cite{akcay2018ganomaly}, which extends with GAN's benefits the DR{\AE}M architecture, with special attention to GAN structure and training loop
				\item The Generative-Reconstructive-Discriminative Network (GRD-Net) architecture, with the attention module based on ROIs.
			\end{enumerate}
		}
		
		\subsection{DR\AE M}
		\label{ch: draem}
		As mentioned before, DR\AE M is an anomaly detection framework based on two different sub-networks. The first subnetwork (called reconstructive sub-network), is trained to recognize anomalies and reconstruct them while keeping the portions of the input image that are not anomalous. The second network learns joint-anomaly inclusion reconstruction to create accurate anomaly segmentation maps by fusing the original and reconstructed appearance.
		
		\medskip
		Instead of generating simulations that accurately reflect the actual appearance of the anomaly in the target domain, DR\AE M instead creates just-out-of-distribution appearances that allow learning the proper distance function to identify the anomaly by its departure from normality. This paradigm is used in the proposed anomaly simulator. The images with artificial anomalies are generated through Perlin noise generator \cite{perlin1985image} to generate a variety of anomaly shapes (see Figure \ref{fig:perlin} (a)). The generated image, is then binarized by a threshold into an anomaly map using uniformly random samples. Then, merging the anomaly map with a random RGB pixels, we obtain the final noise (see Figure \ref{fig:perlin} (b)) to be added to the images of the dataset as can be seen in Figure \ref{fig:perlin} (c). Thus, this process creates training sample triplets with the original image that is free of anomalies, the augmented image that contains simulated anomalies, and the pixel-perfect anomaly mask.
		
		\begin{figure}
			%\centering
			\hspace{-1.5cm}
			\subfloat[Perlin Noise]{\includegraphics[scale=0.126]{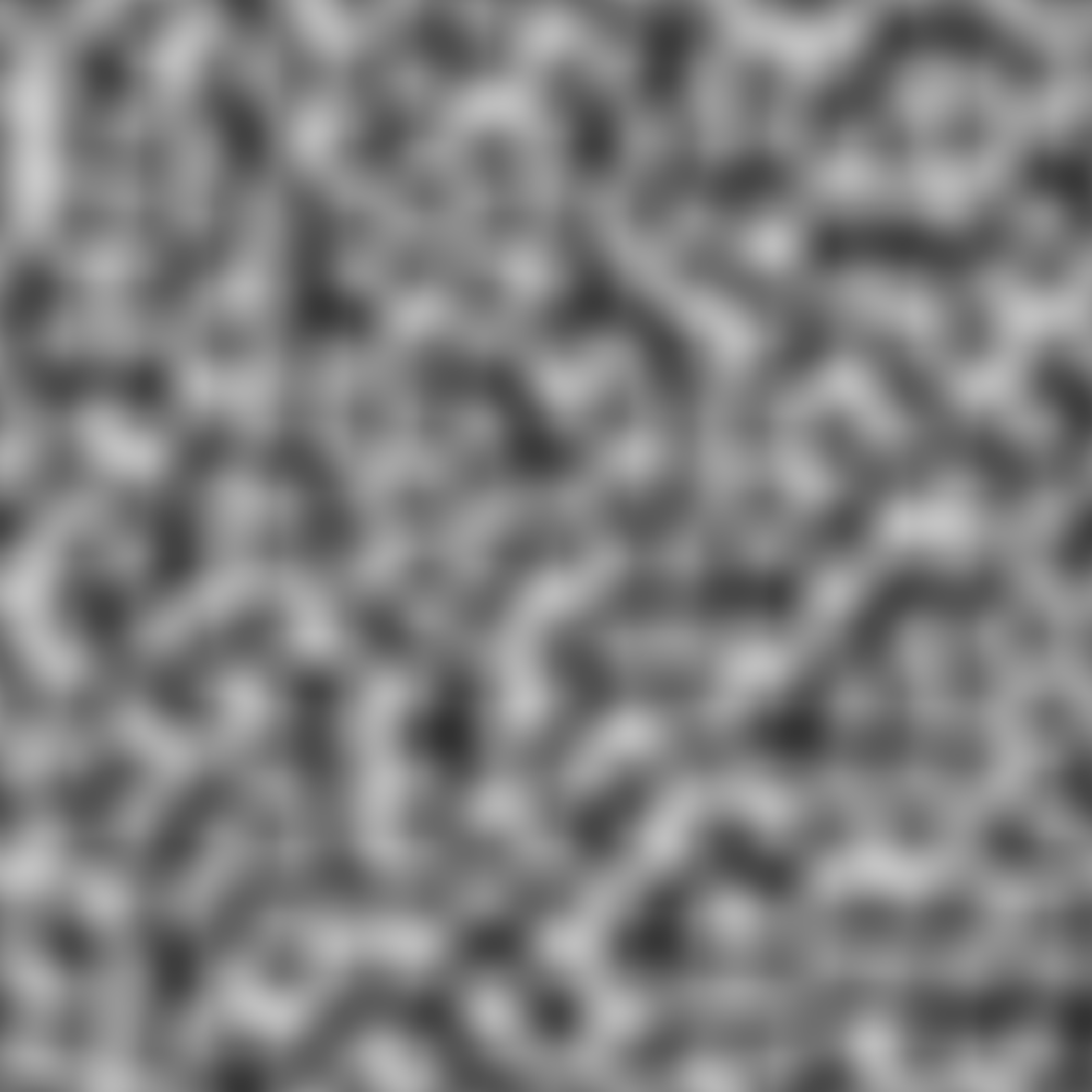}}
			\subfloat[Perlin Noise with random RGB pixels]{\hspace{1.0cm}\includegraphics[scale=0.49]{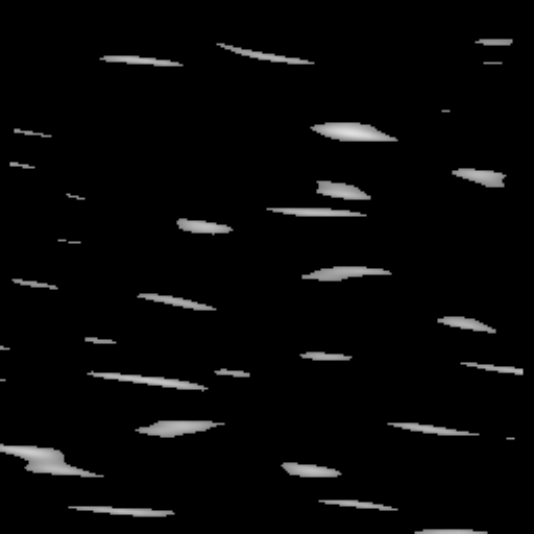}}
			\subfloat[Dataset's image with Perlin noise]{\hspace{1.0cm}\includegraphics[scale=0.49]{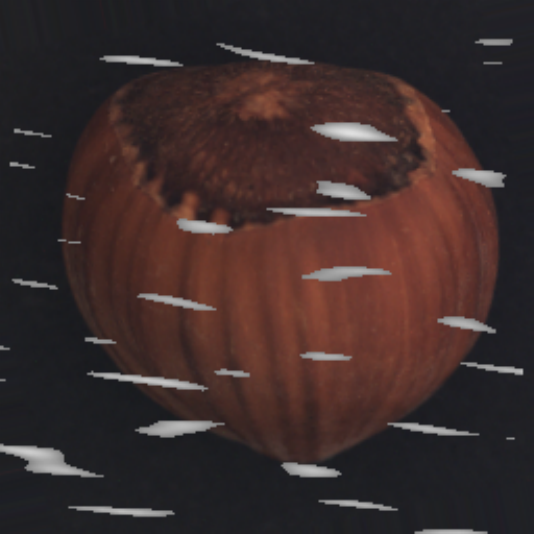}}
			\caption{Simulated anomaly generation process. In (a) there is an example of Perlin noise. (b) represent the merging of the anomaly map and a random RGB pixels. (c) represent and example of a image with generated fake anomalies.}
			\label{fig:perlin}
		\end{figure}
		
		\medskip
		The reconstructive sub-network of DR\AE M perform an image denoising task. It is trained to reconstruct the original image from the artificial corrupted version produced by the process described above. The discriminative sub-network is a U-Net-like neural network that take in input the channel-wise concatenation of the reconstructive sub-net output and the original image. This second sub-network learns to segment the Perlin noise applied to the original image instead to used a similarity functions such as SSIM \cite{wang2004image}.
		
		\medskip
		The output of the discriminative sub-network is an anomaly detection mask. This mask can be interpreted for the image-level anomaly score estimation. The anomaly mask is smoothed by a convolutional filter. The final anomaly score is computed by taking the maximum value of the smoothed anomaly score map.
		
		\subsection{GANomaly}
		\label{ch: ganomaly}
		\subsubsection{Adversarial Autoencoders}
		An Autoencoder (AE) \cite{Goodfellow-et-al-2016} is a neural network that has been trained to attempt to replicate its input to its output. The two components of this network are an encoder (E) that maps the input into latent space $h$ and a decoder (D) that reconstructs the input from the latent space. The ability to constrain $h$ to be smaller than $x$ and the input copying task are where AE's potential lies (in this case, we talk about undercomplete AE). The network is forced to recognise the most crucial aspects of the input data when learning an undercomplete representation. This procedure can be carried out by minimising the the network's penalty function when is far from $x$. To outperform the standard AEs, we can think to train an AE in an adversarial environment \cite{creswell2018generative}. Training AEs with adversarial setting improves reconstruction while also giving the user more control over latent space \cite{makhzani2015adversarial, mirza2014conditional}.
		
		\subsubsection{Generative Adversarial Networks}
		GANs are an unsupervised machine learning approach developed for the task of generate synthetic data \cite{goodfellow2020generative}. Specifically, the first purpose of the GANs was to generate realistic synthetic images. The concept is that during training, two networks - the discriminator and the generator - compete with one another so that the former attempts to generate an image while the latter determines whether it is real or fake. The generator, that is similar to a decoder, learns the distribution of input data from a latent space. 
		
		\subsubsection{GANomaly Architecture \& Training}
		The GANomaly architecture contains two encoders, a decoder, forming an encoder-decoder-encoder structure and discriminator networks \cite{akcay2018ganomaly}. The first encoder-decoder sub-network in an AE that work as the generator part of the model. The generator uses an AE network to reconstruct the input image $x$ after learning how to represent the input data. The second encoder of the encoder-decoder-encoder structure is a network that compress the reconstructed image $\hat{x}$. This encoder has the same architecture on the previous encoder but with different parametrization. This encoder explicitly learns to minimize the distance with its parametrization. This minimization is used during the test to perform anomaly detection. The discriminator network aims to classify the input $x$ and the output $\hat{x}$ as real or fake.
		
		\medskip
		GANomaly is trained by minimizing a loss consisting of three components: the adversarial, contextual and encoder loss. Adversarial loss ($\mathcal{L}_{adv}$) calculated for the discriminator and it is used to reduce the instability of GAN training. Contextual loss ($\mathcal{L}_{con}$) is used to add the contextual information to the final loss. This sub-loss calculates the $\mathcal{L}_{1}$ distance between $x$ and $\hat{x}$, added to $\mathcal{L}_{ssim} = 1 - SSIM$ score loss also calculated between $x$ and $\hat{x}$. So final $\mathcal{L}_{con}$ becomes:
		\begin{equation}
			\label{eq:contextual_loss}
			\mathcal{L}_{con} = \omega_{a}\mathcal{L}_{1}(x, \hat{x}) + \omega_{b}\mathcal{L}_{ssim}(x, \hat{x}).
		\end{equation}
		Finally, the Encoder loss ($\mathcal{L}_{enc}$) is used to minimize the distance between bottleneck features of the input and the encoded features of the generated image. Then, the final loss is describe as:
		\begin{equation}
			\label{eq:ganomaly_loss}
			\mathcal{L}_{gan} = \omega_{1}\mathcal{L}_{adv} + \omega_{2}\mathcal{L}_{con} + \omega_{3}\mathcal{L}_{enc}.
		\end{equation}
		where the weighting parameters ($\omega_{1}$, $\omega_{2}$, and $\omega_{3}$) are used to modify the effect of individual losses on the overall objective function.
		Empirically, it has been found that the best values of the parameters are:
		
		\begin{equation}
			\label{eq:param_omega_values}
			\omega_{a} = 1,	~
			\omega_{b} = 1, ~
			\omega_{1} = 1, ~
			\omega_{2} = 50, ~
			\omega_{3} = 1.
		\end{equation}
		
		These results were obtained starting from the relative reference papers of GANomaly \cite{akcay2018ganomaly}, where $\omega_{1} = 1$, $\omega_{2} = 40$ and $\omega_{3} = 1$ and DR{\AE}M where $\omega_{a} = 1$, $\omega_{b} = 1$. Using a branch and bound approach with a step of $\pm5$ on one $\omega_{*}$ at a time, keeping constant the value of the others. We thus noticed that the weight related to $\mathcal{L}_{con}$, that is $\omega_{2}$, could be increased to $50$ with better result in terms of training time, without losing the contribute of the other components of the main loss.
		
		\subsection{Generative-Reconstructive-Discriminative Network with Attention Module}
		\label{ch: draem_gan}
		
		\begin{figure}
			\centering
			\includegraphics[scale=0.23]{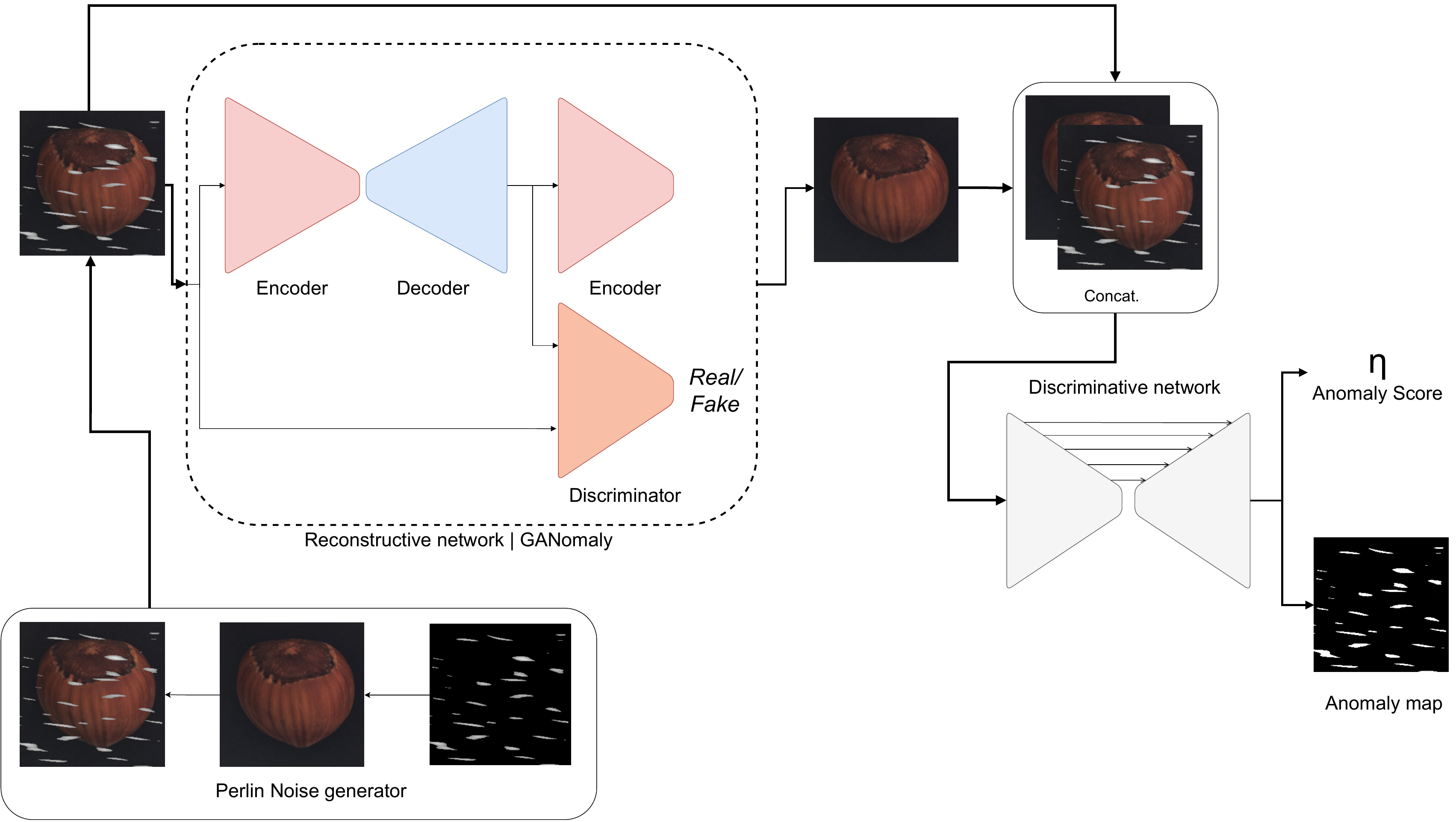}
			\caption{The architecture of DR\AE M GAN. The architecture is quite similar to vanilla DR\AE M, but we can see the implementation of GANomaly instead of the AE which acted as the Reconstructive network.}
			\label{fig:architecture}
		\end{figure}
		
		This work is heavily inspired by DR\AE M \cite{zavrtanik2021draem}. This is reflected in the general architecture of the proposed framework. As you can see in Figure \ref{fig:architecture}, the architecture is quite similar to the vanilla DR\AE M but with the difference that instead of the AE which acted as Reconstructive network, there is now an implementation of GANomaly. All networks engaged in the reconstructive sub-network are residual to avoid degradation problems during the training.
		Then, to train the GANomaly engaged in GRD-Net, the loss described in Equation \ref {eq:ganomaly_loss} is used. To train the discriminative network, Focal Loss (FL) \cite{lin2017focal, zavrtanik2021draem} is used. FL can be defined by the equation:
		
		\begin{equation}
				\mathcal{FL}(p) = -(1 - p)^{\gamma}\log(p).
		\end{equation}
		
		Basically, $\mathcal{FL}$ adds the factor $-(1 - p)^{\gamma}$ to the standard cross entropy. Setting $\gamma > 0$ reduces the relative loss for the well classified images, putting more focus on the misclassified examples \cite{lin2017focal}.
		This loss applied on this sub-network, increase robustness towards accurate segmentation of hard examples. A further improvement was applied to the discriminatory network. In order to ensure that only the defects present on the surface of the inspected products are considered. To do this, in addition to the images of the dataset, the network is also given a segmentation mask that highlights the area of interest (AOI) of the product. This mask is multiplied by the anomaly detection mask to obtain an \emph{intersection mask}. Then, $\mathcal{FL}$ is calculated on this intersection. 
		The overall loss of GRD-Net became:
		\begin{equation}
				\mathcal{I} = \mathcal{A}_{discr} \times \mathcal{ROI}_{input}.
				\label{eq:int_eq}
		\end{equation}
		
		Where $\mathcal{I}$ is the \textit{intersection} mask, that is a tensor obtained by the intersection (multiplication) of the input mask tensor $\mathcal{ROI}_{input}$, that highlight a ROI (Region Of Interest) in which the network has to segment the anomaly area, and the output mask tensor $\mathcal{A}_{discr}$ of the discriminative network, that segment the original image
			
		\begin{equation}
			\mathcal{L}_{tot} = \mathcal{L}_{gan} + \mathcal{FL}(\mathcal{I}, \mathcal{M}_{input}).
			\label{eq:def_floss}
		\end{equation}
		
		So the total loss function $\mathcal{L}_{tot}$ is the sum of the GAN Loss $\mathcal{L}_{gan}$ and the Focal Loss calculated on the intersection area $\mathcal{FL}(\mathcal{I})$
		
		{
			\begin{figure}
				\centering
				\includegraphics[scale=1]{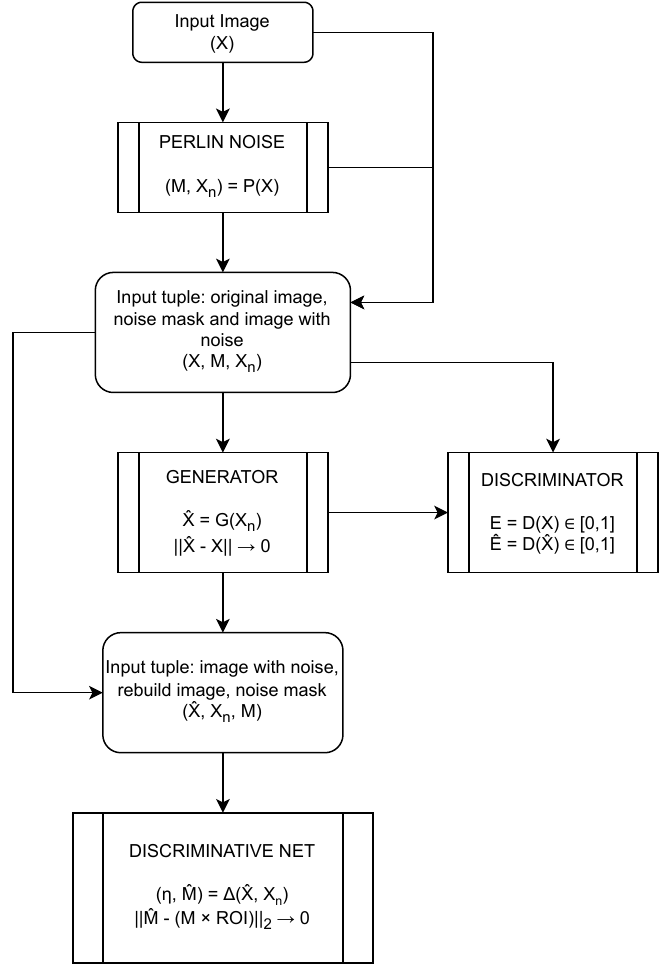}
				\caption{Train step flowchart: input image $X$ is transformed in $X_n$, that is the image with the Perlin noise superimposed. $M$ is the mask image of the noise areas.}
				\label{fig:train_flow}
			\end{figure}
			
			\begin{figure}
				\centering
				\includegraphics[scale=1]{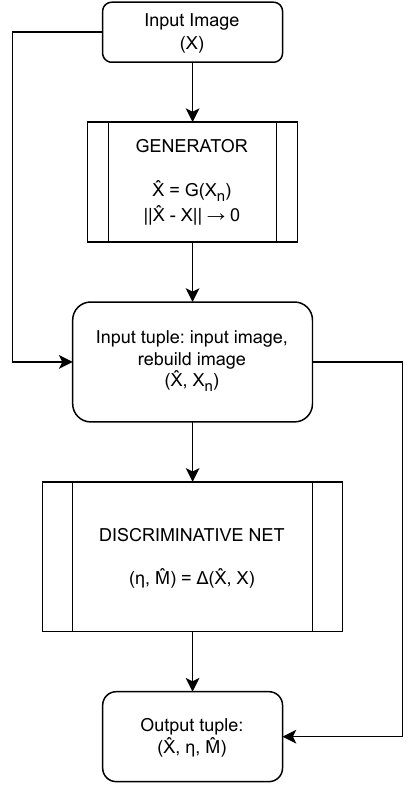}
				\caption{Inference step flowchart.}
				\label{fig:inference_flow}
			\end{figure}
		}
		
		Finally, the overall training and inference sequences are schematized respectively in Figures {\ref{fig:train_flow}} and {\ref{fig:inference_flow}}.
	}

\section{Experiments}
	\label{sec: experiments}
	{
		Several experiments were performed to test the performance of  GRD-Net.
		First of all the performances of the GAN with residual convolutional autoencoder have been compared and evaluated with DR{\AE}M and GANomaly, which represent state-of-the-art reconstruction-based anomaly detection and localization technologies. A schema of one stage composed of two residual blocks is shown in Figure \ref{fig:netarch}.
		
		\medskip
		Experiments were conducted over 200 training epochs using vanilla DR{\AE}M autoencoder and our GAN with residual AE.
		Our network takes inspiration from the ResNet V2 architecture used for the classification task \cite{DBLP:journals/corr/HeZRS15}.
		
		\medskip
		{
			\medskip
			In this section, summarizing:
			\begin{enumerate}
				\item The residual block applied to our architecture
				\item The two phases of the training loop for the generative part and the discriminative part
				\item The benefits of residual network applied to autoencoder of generative part of the GAN.
				\item Experiments on the Generative-Reconstructive-Discriminative Network (GRD-Net) architecture, with the attention module based on ROIs.
				\item A real-case experiments based on pharmaceutical BFS vials, with attention on the body of the aforementioned vials.
			\end{enumerate}
		}
		
		\begin{center}
			\begin{figure}
				\centering
				%\hspace{-1.5cm}
				\includegraphics[scale=0.75]{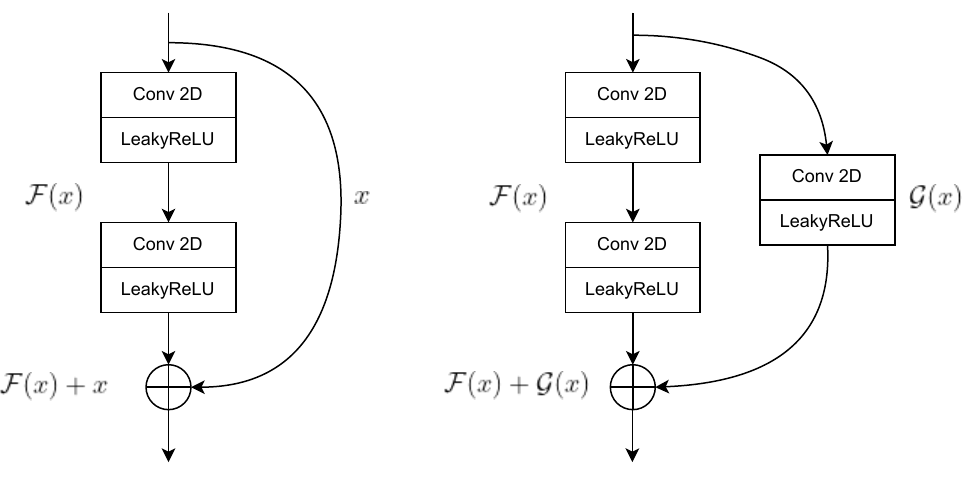}
				\caption{Two consecutive residual blocks of one stage of the encoder network. The introduction of a residual architecture in the encoder-decoder-encoder GAN \cite{DBLP:journals/corr/HeZRS15} revealed to be more stable during training phase, by giving better results with equal epochs.}
				\label{fig:netarch}
			\end{figure}
		\end{center}
		
		The first part of the experiments was conducted using three challenging datasets from MVTec's sets: hazelnut, metal nut and pill datasets.
		In the second part, the network was tested on hazelnut, zip and a proprietary pharmaceutical set of BFS strips of vials, from a real study and use case that took place in Bonfiglioli Engineering, for a quality control vision inspection machine.
		For those datasets, a second ROI dataset was prepared for each training nominal image.
		
		\subsection{GAN with Residual AE}
		As mentioned above, the first experiment aims to challenge the vanilla version of DR{\AE}M architecture.
		The training lasted 200 epochs, instead of the 700 used for testing DR{\AE}M on the original paper, this to provide a more realistic case, which can be implemented on a production line in a real industrial field.
		During this step were evaluated anomaly detection per image and defect localization within the image.
		Learning rate is set to $10^{-4}$ and we used a policy based on {\textquotedblleft}\textit{reduction on plateau}{\textquotedblright} heuristic with a patience of 3 epochs and a reduction factor $\alpha = 0.1$. When a plateau of 3 epochs is reached at epoch $k$ it decreases using the formula:
		\begin{equation}
			\mathcal{LR}_{k} = \mathcal{LR}_{k-1} \cdot e^{-\alpha}.
			\label{learningrate}
		\end{equation}
		
		Where $\mathcal{LR}_{k}$ is the learning rate at the k-th epoch.
		
		For the evaluation, we used the AUROC, widely used in architecture comparisons, at image-level and at pixel-level, as semi-supervised anomaly detection and localization score.
		
		Data augmentation is performed on training examples, using a random rotation in the range of $[-{{\pi} \over {2}}, +{{{\pi} \over {2}}}]$ radians, in order to reduce overfitting during training over lots of epochs, because of the small number of anomaly-free images provided in MVTec datasets.
		
		\begin{table}[h]
			\begin{center} {\footnotesize
					\begin{tabular}{lcc}
						\hline
						& \multicolumn{2}{c}{AUROC per image (pixel) at 10 epochs}  	\\
						& \multicolumn{1}{c}{DR\AE M}
						& \multicolumn{1}{c}{GRD-Net}			\\
						\hline
						& 73.1 & 96.7 \\[-2ex]
						\raisebox{2ex}{hazelnut}
						& (55.7) & (91.0)\\[0ex]
						& 58.3 & 96.4 \\[-2ex]
						\raisebox{2ex}{metal nut}
						& (49.0) & (69.3)\\[0ex]
						& 74.2 & 77.5 \\[-2ex]
						\raisebox{2ex}{pill}
						& (66.0) & (90.5)\\[0ex]
						\hline
				\end{tabular} }
			\end{center}
			\caption{\footnotesize AUROC score after 10 epochs of training per image and (pixel).}
			\label{table:auroc020}
		\end{table}
		
		\begin{table}[h]
			\begin{center} {\footnotesize
					\begin{tabular}{lcc}
						\hline
						& \multicolumn{2}{c}{AUROC per image (pixel) at 35 epochs}  	\\
						& \multicolumn{1}{c}{DR\AE M}
						& \multicolumn{1}{c}{GRD-NET}			\\
						\hline
						& 85.3 & 99.5 \\[-2ex]
						\raisebox{2ex}{hazelnut}
						& (82.4) & (95.5)\\[0ex]
						& 61.8 & 99.3 \\[-2ex]
						\raisebox{2ex}{metal nut}
						& (49.0) & (69.3)\\[0ex]
						& 75.7 & 89.8 \\[-2ex]
						\raisebox{2ex}{pill}
						& (86.5) & (95.5)\\[0ex]
						\hline
				\end{tabular} }
			\end{center}
			\caption{\footnotesize AUROC score after 35 epochs of training per image and (pixel).}
			\label{table:auroc050}
		\end{table}
		
		\begin{table}[h]
			\begin{center} {\footnotesize
					\begin{tabular}{lcc}
						\hline
						& \multicolumn{2}{c}{AUROC per image (pixel) at 100 epochs}  	\\
						& \multicolumn{1}{c}{DR\AE M}
						& \multicolumn{1}{c}{GRD-Net}			\\
						\hline
						& 98.8 & 100.0 \\[-2ex]
						\raisebox{2ex}{hazelnut}
						& (94.8) & (97.3)\\[0ex]
						& 99.7 & 99.8 \\[-2ex]
						\raisebox{2ex}{metal nut}
						& (86.7) & (70.4)\\[0ex]
						& 93.8 & 98.2 \\[-2ex]
						\raisebox{2ex}{pill}
						& (94.8) & (95.5)\\[0ex]
						\hline
				\end{tabular} }
			\end{center}
			\caption{\footnotesize AUROC score after 100 epochs of training per image and (pixel).}
			\label{table:auroc100}
		\end{table}
		
		\begin{table}[h]
			\begin{center} {\footnotesize
					\begin{tabular}{lccccc}
						\hline
						& \multicolumn{3}{c}{AUROC per image (pixel)}  	\\
						& \multicolumn{1}{c}{GANomaly}
						& \multicolumn{1}{c}{DR\AE M}
						& \multicolumn{1}{c}{PaDiM}
						& \multicolumn{1}{c}{PatchCore}
						& \multicolumn{1}{c}{GRD-Net}			\\
						\hline
						& 78.5 & 100.0 & {-} & {100.0} & 100.0 \\[-2ex]
						\raisebox{2ex}{hazelnut}
						& (-) & (95.0) & {97.7} & {98.6} & (97.4)\\[0ex]
						& 70.0 & 98.7 & {-} & {99.7} & 100.0 \\[-2ex]
						\raisebox{2ex}{metal nut}
						& (-) & (86.7) & {96.7} & {98.4} & (96.2)\\[0ex]
						& 74.3 & 97.9 & {-} & {97.0} & 98.5 \\[-2ex]
						\raisebox{2ex}{pill}
						& (-) & (94.8) & {94.7} & {97.1} & (95.8)\\[0ex]
						& {75.7} & {91.8} & {-} & {99.3} & {99.5} \\[-2ex]
						{\raisebox{2ex}{cable}}
						& {(-)} & {(94.7)} & {96.7} & {98.2} & {(98.1)}\\[0ex]
						\hline
				\end{tabular} }
			\end{center}
			\caption{\footnotesize Final comparative table with AUROC score between GANomaly (200 epochs), DR{\AE}M (200 epochs), PaDiM (ResNet18 pre-train), PatchCore (ResNet50 pre-train) and GRD-Net (200 epochs). The results of GANomaly and DR{\AE}M. are obtained by us adjusting the number of training epochs to the number of training epochs used to train GRD-Net, for this reason the final result may vary a little from the reference paper.}
			\label{table:auroc200}
		\end{table}
		
		{
			For the sake of completeness, we also tested and compared GRD-Net with a vanilla convolutional autoencoder (that is without residual block), and the GRD-Net with fully-convolutional residual autoencoder.
			
			The experiment was performed using our huge pharmaceutical dataset on 500 epochs, using only the generative part, comparing the losses. This since the second part depends strictly on the performance of the first.
			
			The experimental results are very encouraging in support of the intuition that the residual network, even in the case of an autoencoder applied to a GAN, is more effective in generating the final data.
			
			\begin{figure}
				\centering
				%\hspace{-1.5cm}
				\subfloat[Training adversarial loss (magenta for vanilla and cyan for residual)]{\includegraphics[scale=0.6]{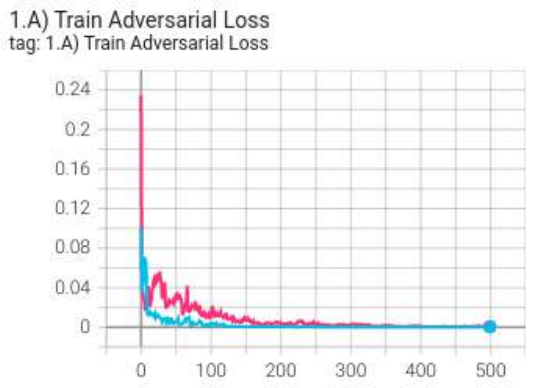}}
				\hspace{1.0cm}\subfloat[Training contextual loss (magenta for vanilla and cyan for residual)]{\includegraphics[scale=0.6]{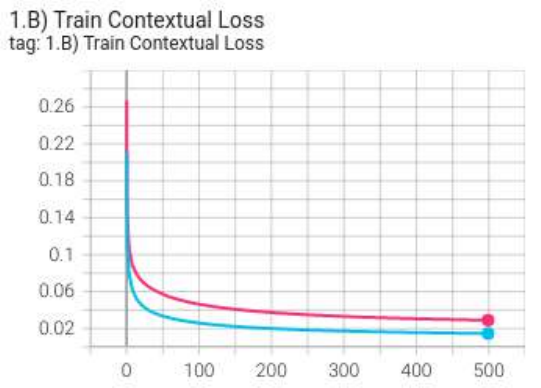}}
				\hspace{1.0cm}\subfloat[Training encoder loss (magenta for vanilla and cyan for residual)]{\includegraphics[scale=0.6]{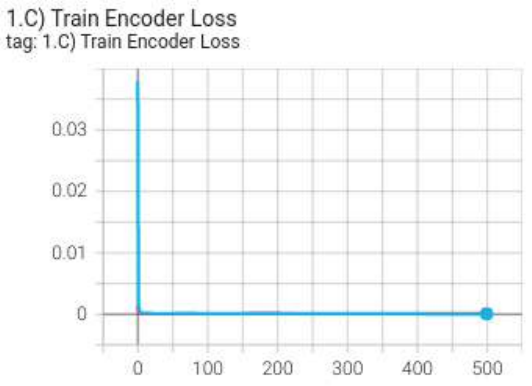}}
				\hspace{1.0cm}\subfloat[Training SSIM loss (magenta for vanilla and cyan for residual)]{\includegraphics[scale=0.6]{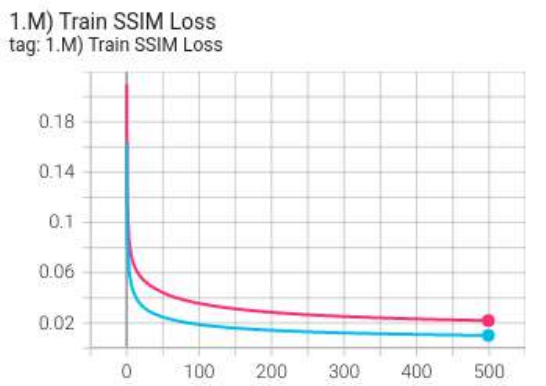}}
				\hspace{1.0cm}\subfloat[Validation adversarial loss (magenta for vanilla and cyan for residual)]{\includegraphics[scale=0.6]{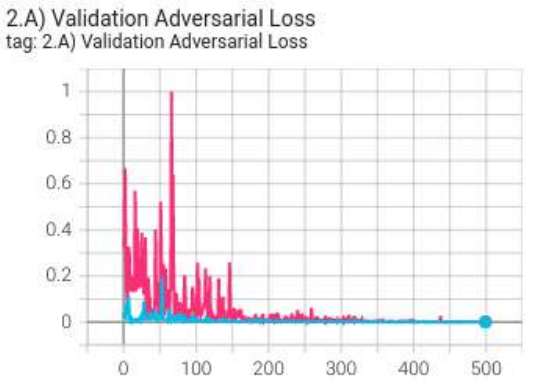}}
				\hspace{1.0cm}\subfloat[Validation contextual loss (magenta for vanilla and cyan for residual)]{\includegraphics[scale=0.6]{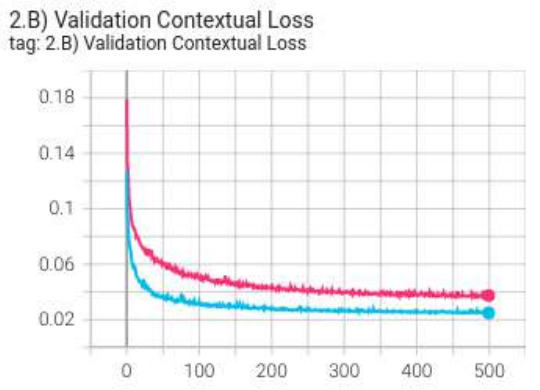}}
				\hspace{1.0cm}\subfloat[Validation encoder loss (magenta for vanilla and cyan for residual)]{\includegraphics[scale=0.6]{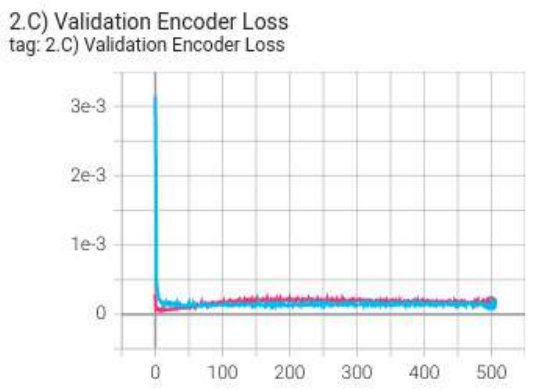}}
				\hspace{1.0cm}\subfloat[Validation SSIM loss (magenta for vanilla and cyan for residual)]{\includegraphics[scale=0.6]{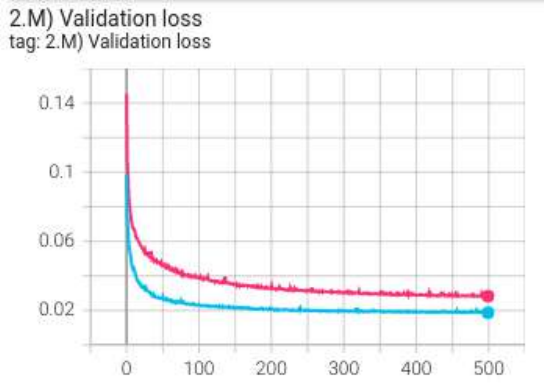}}
				\caption{Visual representation of how the network with vanilla autoencoder (magenta) is not only less effective, but also noisier in some losses, such as adversarial loss, compared to the residual architecture (cyan).}
				\label{fig:losses_comparison}
			\end{figure}
			
			\begin{table}[h]
				\begin{center} {\footnotesize
						\begin{tabular}{lcccc}
							\hline
							& \multicolumn{3}{c}{Loss value for training (validation) phase}  	\\
							& \multicolumn{1}{c}{Epoch 250 vanilla}
							& \multicolumn{1}{c}{Epoch 250 residual}
							& \multicolumn{1}{c}{Epoch 500 vanilla}
							& \multicolumn{1}{c}{Epoch 500 residual}			\\
							\hline
							& $2.8811 \times 10^{-3}$ & $3.2498 \times 10^{-4}$ & $7.4750 \times 10^{-4}$ & $1.3797 \times 10^{-4}$ \\[-2ex]
							\raisebox{2ex}{Adversarial Loss}
							& ($5.1180 \times 10^{-2}$) & ($3.8928 \times 10^{-3}$) & ($1.3720 \times 10^{-3}$) & ($4.5011 \times 10^{-4}$) \\[0ex]
							& 0.03502 & 0.01853 & 0.02912 & 0.01460 \\[-2ex]
							\raisebox{2ex}{Contextual loss}
							& (0.04136) & (0.02782) & (0.03747) & (0.02488) \\[0ex]
							& $1.7221 \times 10^{-4}$ & $6.8905 \times 10^{-5}$ & $1.1745 \times 10^{-4}$ & $4.6982 \times 10^{-5}$ \\[-2ex]
							\raisebox{2ex}{Encoder Loss}
							& ($2.1078 \times 10^{-4}$) & ($1.2333 \times 10^{-4}$) & ($1.6965 \times 10^{-4}$) & ($1.4363 \times 10^{-4}$) \\[0ex]
							& 0.02665 & 0.01304 & 0.02200 & 0.01014 \\[-2ex]
							\raisebox{2ex}{SSIM Loss}
							& (0.03115) & (0.02010) & (0.02830) & (0.01872) \\[0ex]
							\hline
					\end{tabular} }
				\end{center}
				\caption{{Comparison between \textit{vanilla} and \textit{residual} architecture used for generative part of the GAN architecture in the GRD-Net.}}
				\label{table:loss_comparison}
			\end{table}
			
			This can be visually appreciated in the Figure \ref{fig:losses_comparison}. We also provided a comparison between losses used for the generator in Table \ref{table:loss_comparison}.
			
		}
		
		\paragraph{Anomaly Detection}
		For what concerns surface anomaly detection, our proposed architecture enhance somewhat not only the final score of the two reference models but improves also the learning curve making it smoother and steeper toward convergence, especially during the first transitional period.
		In addition to this, also the difference between training and validation curves is far less with our model.
		The smoothing of the learning curve can be explained by the GAN model that, with the discriminator network, improve the stability of the training process.
		The steepest incline and a lower presence of the overfitting phenomenon (that can be observed with the higher difference between validation and training curves), can be attributed to both the GAN model and residual net.
		This is due to the improvement given by the adversarial part of the GAN and by the reduction of the gradient vanishing that could affect deep convolutional networks.
		
		\paragraph{Anomaly Localization}
		As for anomaly detection, also anomaly localization has been compared with DR\AE M after 200 epochs of training.
		GANomaly was not included in this comparison because does not exist, in the official paper, a method capable of locating defective regions.
		Table \ref{table:auroc020}, \ref{table:auroc050}, \ref{table:auroc100} and Table \ref{table:auroc200} show the AUROC result comparison between DR\AE M and our approach, as mentioned above in four different stages of the training phase: after 10, 50, 100 and 200 epochs.
		The results are very encouraging because they improve those of the vanilla network, in fact, a better quality of the reconstructed image implies a better performance of the second \textit{discriminative} network.
		Moreover, as in Figure \ref{fig:losses}, it's clear how validation curves are much better in our (red) model, compared to the vanilla one (orange), especially in the first phase of the training; reducing, thus, the number of needed epochs for obtaining an acceptable result for an industrial process.
		
		On the other hand, \emph{embedding similarity-based} network, like PatchCore, seems to have a better pixelwise AUROC score. But because of the nature itself of the architecture, it is not possible to add, in an easy way, an attention module based on ROIs.
		
		\begin{figure}
			\centering
			%\hspace{-1.5cm}
			\subfloat[Validation contextual loss]{\includegraphics[scale=0.6]{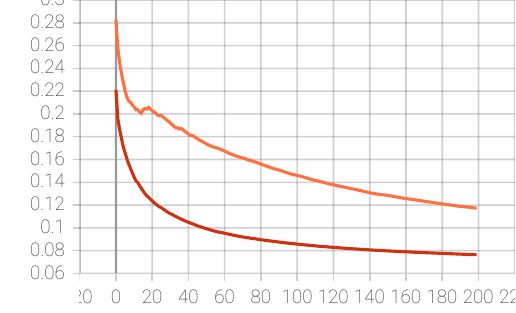}}
			\hspace{1.0cm}\subfloat[Validation Focal Loss]{\includegraphics[scale=0.6]{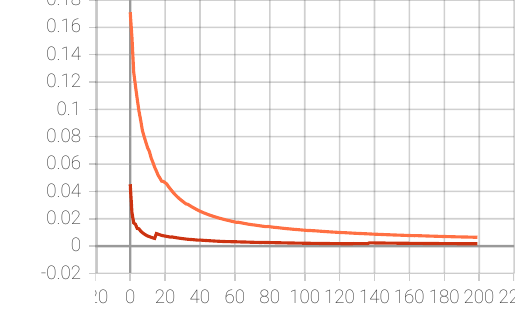}}
			\caption{Validation losses for generative (a) and discriminative (b) sub-networks. Red curve is obtained during training of our model, the orange one is obtained with the vanilla model. It is evident that the learning curve is much better in our case, for both nets.}
			\label{fig:losses}
		\end{figure}
		
		\subsection{GRD-Net with ROI}
		In the second experiment was tested the capability of learning an interest region within the image in which and only in which spot and locate the anomalies.
		Zipper and Hazelnut datasets were used for the purpose. Zipper, especially, is particularly suitable, since samples have 2 logic regions of interest: the zipper area itself and the fabric area.
		In our case we used as region of interest the zipper part, so we would exclude defects on the fabric zone. An example is showed in Figure \ref{fig:results}.
		As previously explained in Section \ref{ch: draem_gan}, \textit{discriminative} network was trained using as focal loss variable intersection between ROI and the mask generated from Perlin noise.
		In this way, the net will start generalizing not only how to spot anomalies from differences between original and reconstructed images, but also where is located the area in which look for differences.
		In fact, in most industrial cases, the totality of the image is not important; indeed, sometimes it could be misleading, as there may be anomalies within the frame that are not part of the product itself.
		
		\begin{figure}
			\centering
			%\hspace{-1.5cm}
			\subfloat[Original image ($X$)]{\includegraphics[scale=0.5]{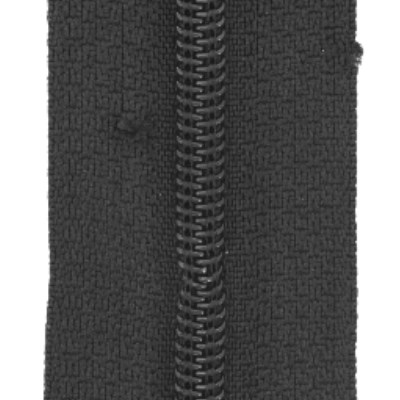}}
			\hspace{0.5cm}\subfloat[Reconstructed image by the Generator $G$ ($\hat{X}$)]{\includegraphics[scale=0.5]{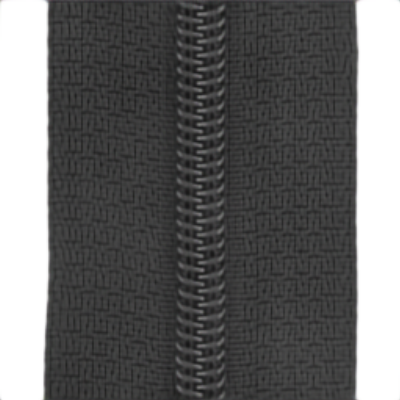}}
			\hspace{0.5cm}\subfloat[Ground truth ($M$)]{\includegraphics[scale=0.5]{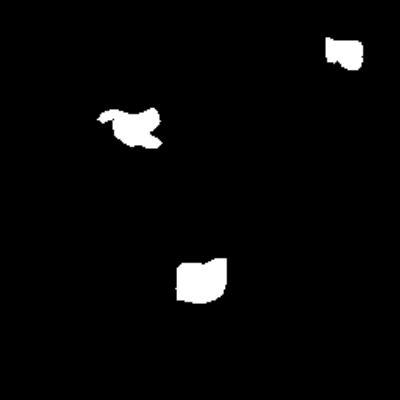}}
			\hspace{0.5cm}\subfloat[Generated \textit{heatmap} by \textit{discriminative} model]{\includegraphics[scale=0.5]{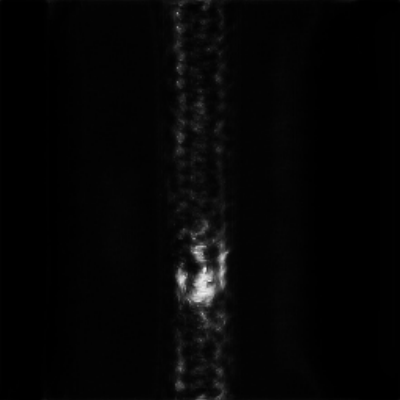}}
			\hspace{0.5cm}\subfloat[Generated \textit{heatmap} by \textit{discriminative} model after average pooling with $21 \times 21$ kernel]{\includegraphics[scale=0.5]{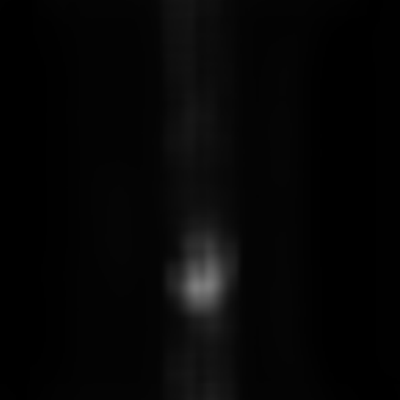}}
			\hspace{1.0cm}\subfloat[Result generated anomaly localization region ($\hat{M}$)]{\includegraphics[scale=0.5]{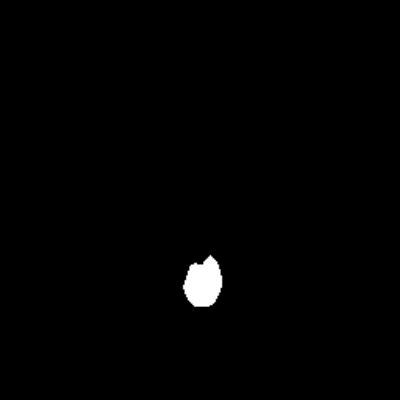}}
			\hspace{0.5cm}\subfloat[Original image with regions: \textbf{blue} region is the \textit{ROI}; the \textbf{orange} region is the ground truth ($M$); and finally the \textbf{red} region is the generated region ($\hat{M}$)]{\includegraphics[scale=0.2]{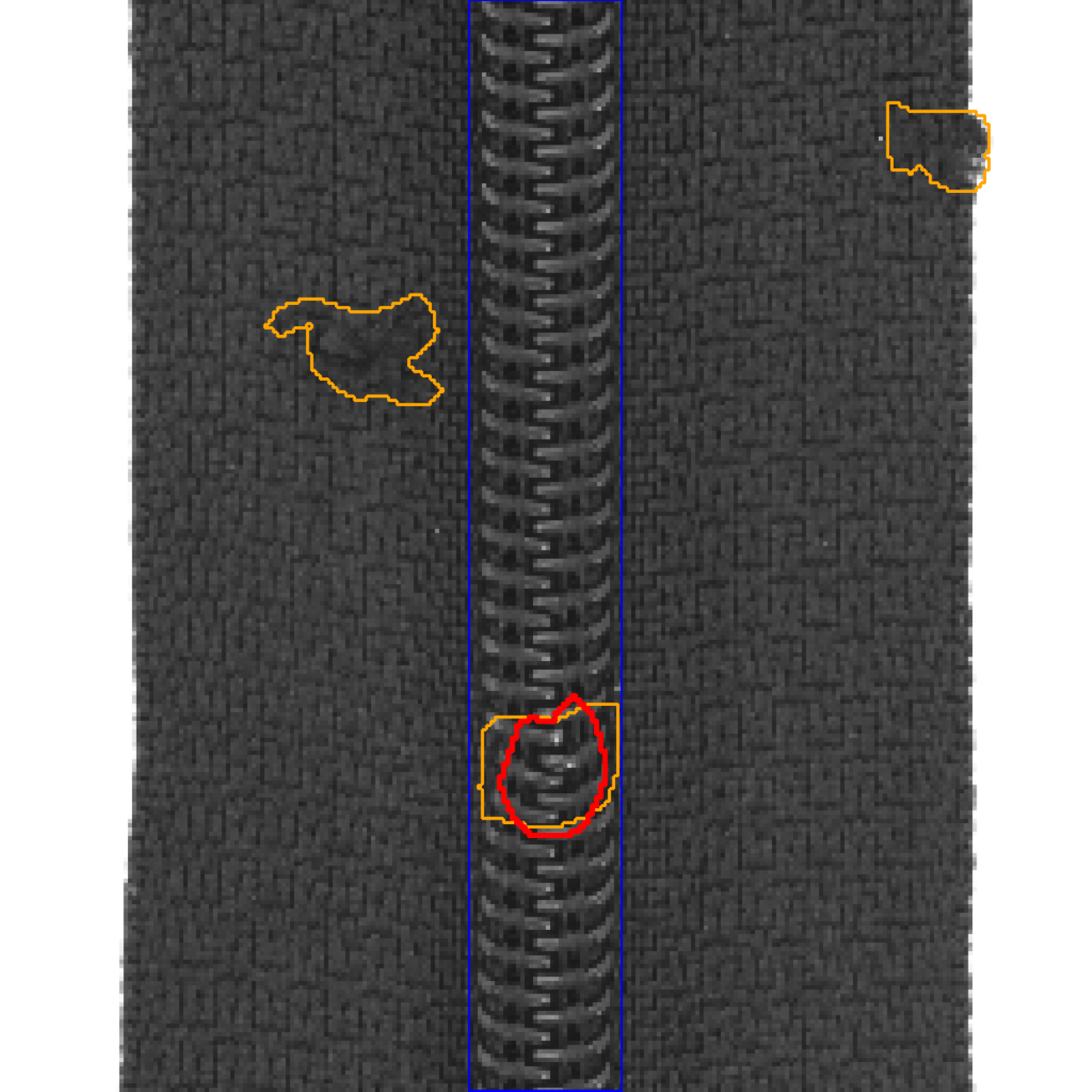}}
			\hspace{0.5cm}\subfloat[Image generated overimposing the convoluted heatmap from the \textit{discriminative} net to $X$, and colorizing it with \textit{jet} color-map]{\includegraphics[scale=0.2]{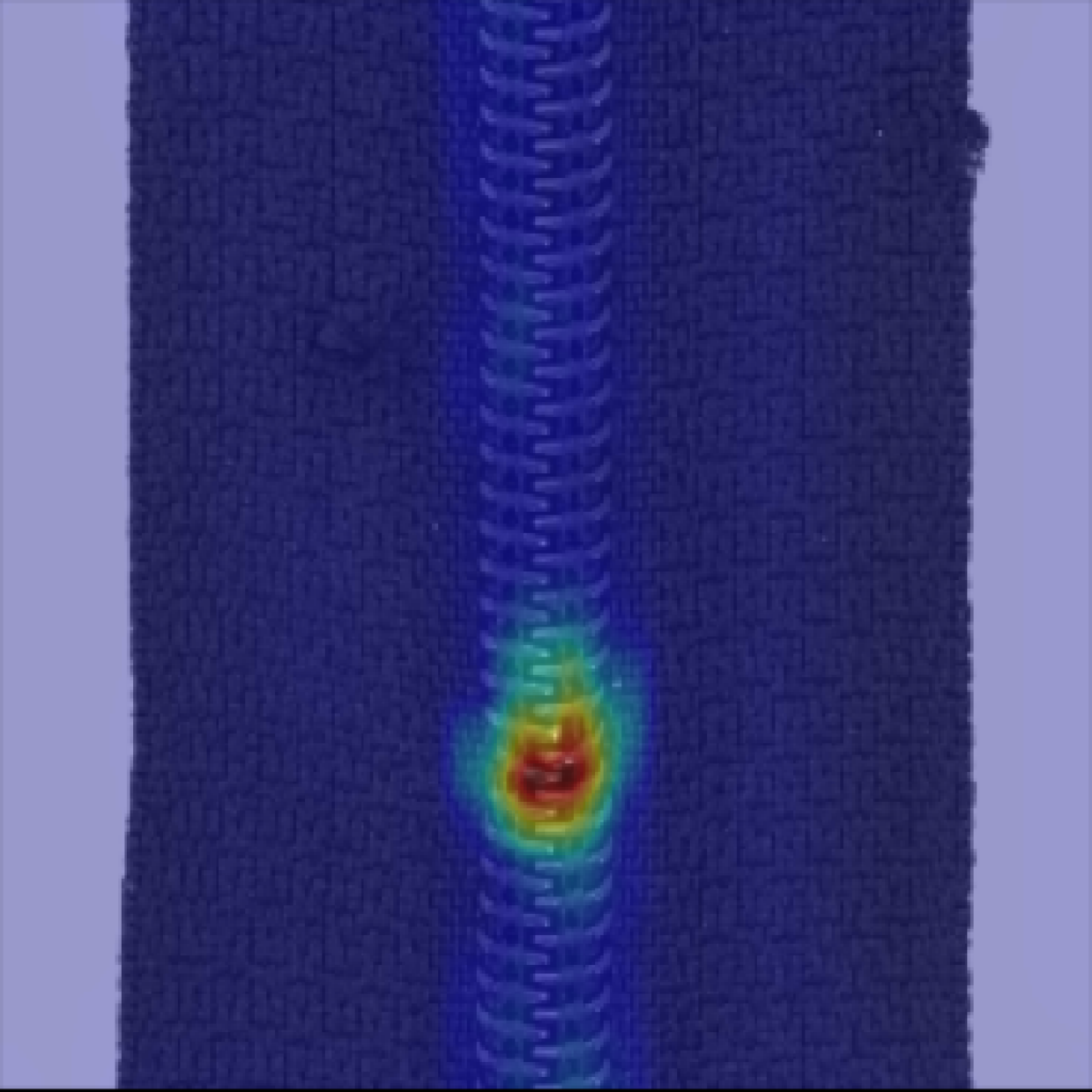}}
			\caption{Especially significant example from zipper dataset in which we could spot 3 anomalies: one in the zipper, one in the middle of fabric part and another, the last, on the border of the fabric zone. As we can see, the only spotted is the one in the zipper region, perfectly inside ROI, and almost perfectly aligned with the ground truth defect region.}
			\label{fig:results}
		\end{figure}
		
		In order to obtain this result, were created a ROI for each training image and the focal loss was customized as explained in Paragraph \ref{ch: draem_gan}, namely by intersecting mask during training and the aforementioned region of interest.
		Thus, the \textit{discriminative} network learn to generalize the most important part of the image, where to focus the attention.
		
		\subsection{Real-Case Experiment}
		The studied model was used in a real-case industrial process to perform a quality control on pharmaceutical BFS strips of vials.
		Tests are performed by a Bonfiglioli Engineering automatic machine, with a rotary carousel with a tracker where are installed acquiring sensors.
		Training set is composed by 230355 images of vials, acquired in 3 different areas by a on-line camera, during production process.
		For reasons related to non-disclosure agreements, we cannot show full product images, but only a limited area, that cover one of most interesting part to our aim.
		Strips consist of 5 plastic material, namely BFS, vials, stick to each other on the side, liquid filled.
		Because of these features, one of the most challenging areas is the \textit{meniscus} region. This due to the great randomness and variability of the aforementioned meniscus. Its own shape and the possibility that there could be bubbles under it or liquid drops over it, make very difficult to treat this region using only classical blob-analysis algorithms.
		
		\begin{figure}
			\centering
			%\hspace{-1.5cm}
			
			\subfloat[{Floating black particle on meniscus, near the \textit{shoulder} of the vial}]{\includegraphics[scale=0.5]{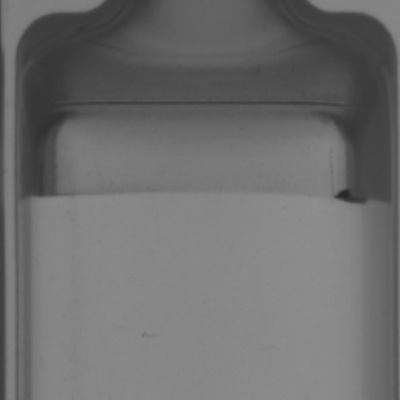}}
			\hspace{0.25cm}\subfloat[Black spot near the meniscus]{\includegraphics[scale=0.5]{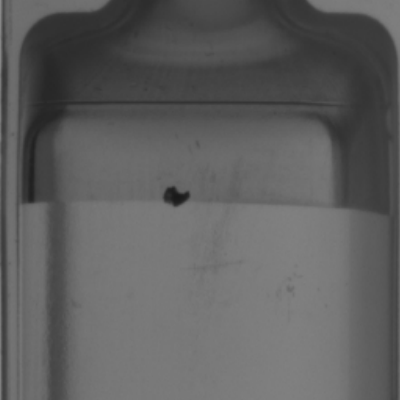}}
			\hspace{0.25cm}\subfloat[{Scratch at the turn of horizontal engraving}]{\includegraphics[scale=0.5]{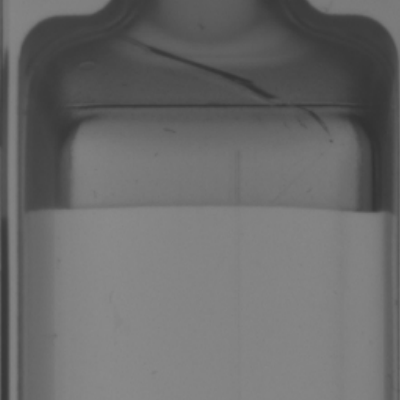}}
			
			\caption{3 examples of real cases where algorithmic analysis is difficult, if not almost impossible.}
			\label{fig:real_case_examples}
		\end{figure}
		
		In Figure \ref{fig:real_case_examples} are shown 3 real-case examples where blob-analysis is almost impossible due to the variability of the meniscus shape and the shadows generated by the shape of the product itself and the position of the sensor in relation of the product.
		
		\begin{figure}
			\centering
			%\hspace{-1.5cm}
			
			\raisebox{6\height}{Case 1:}
			\hspace{0.25cm}\subfloat[Original image $X$ with dark floating particle on meniscus]{\includegraphics[scale=0.5]{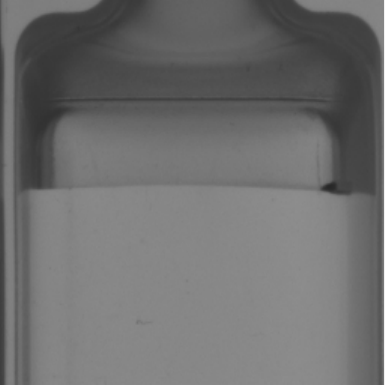}}
			\hspace{0.25cm}\subfloat[Generated heatmap $\hat{M}$ normalized between 0 and 1]{\includegraphics[scale=0.5]{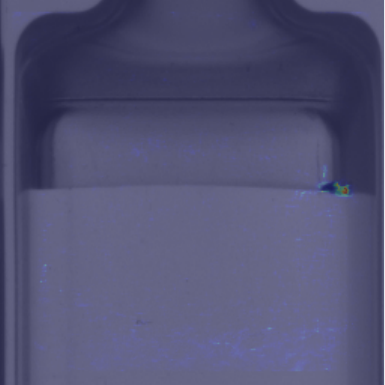}}
			\hspace{0.25cm}\subfloat[Defect localization after convolution and threshold]{\includegraphics[scale=0.5]{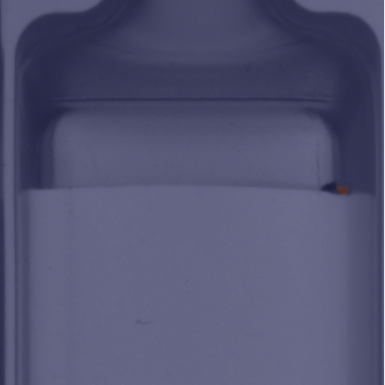}}
			
			\raisebox{6\height}{Case 2:}
			\hspace{0.25cm}\subfloat[Original image $X$ with black spot on vial surface]{\includegraphics[scale=0.5]{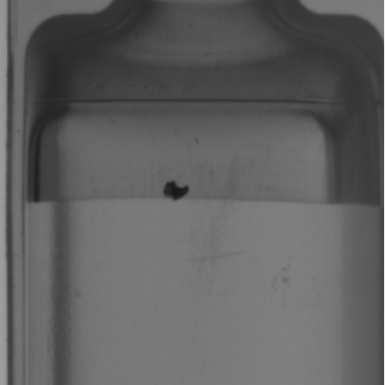}}
			\hspace{0.25cm}\subfloat[Generated heatmap $\hat{M}$ normalized between 0 and 1]{\includegraphics[scale=0.5]{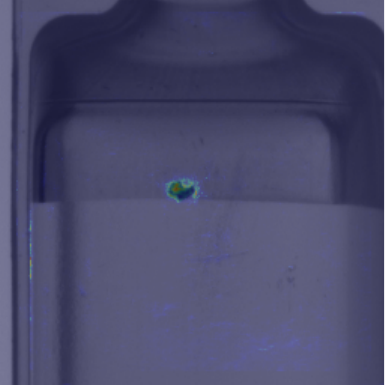}}
			\hspace{0.25cm}\subfloat[Defect localization after convolution and threshold]{\includegraphics[scale=0.5]{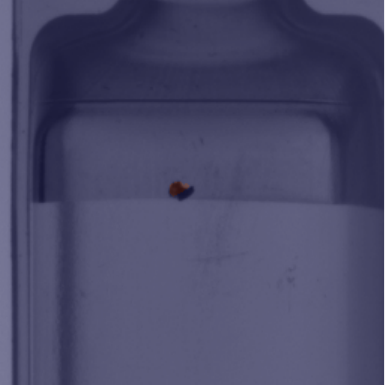}}
			
			\raisebox{6\height}{Case 3:}
			\hspace{0.25cm}\subfloat[Original image $X$]{\includegraphics[scale=0.5]{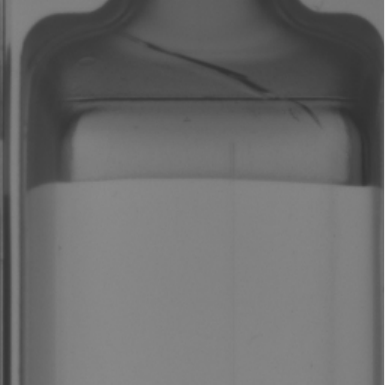}}
			\hspace{0.25cm}\subfloat[Generated heatmap $\hat{M}$ normalized between 0 and 1]{\includegraphics[scale=0.5]{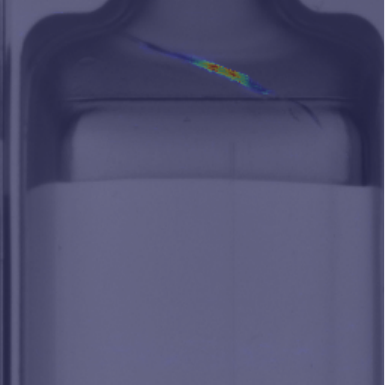}}
			\hspace{0.25cm}\subfloat[Defect localization after convolution and threshold]{\includegraphics[scale=0.5]{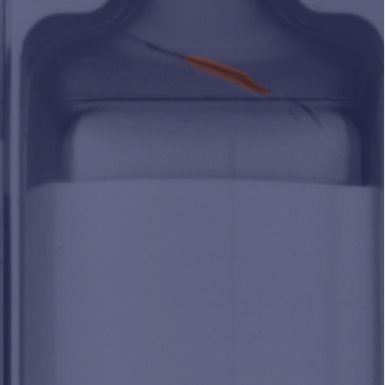}}
			
			\raisebox{6\height}{Case 4:}
			\hspace{0.25cm}\subfloat[Original image $X$ without defect but a system of bubble near meniscus]{\includegraphics[scale=0.5]{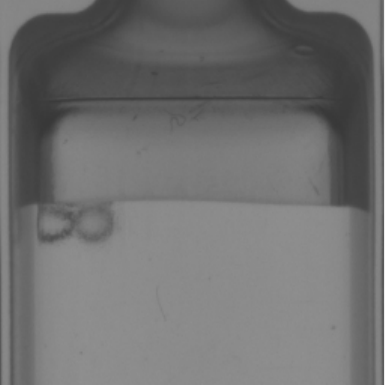}}
			\hspace{0.25cm}\subfloat[Generated heatmap $\hat{M}$ normalized between 0 and 1]{\includegraphics[scale=0.5]{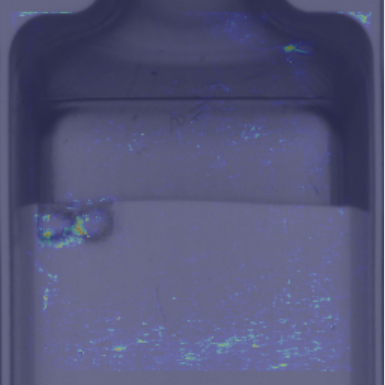}}
			\hspace{0.25cm}\subfloat[Defect localization after convolution and threshold]{\includegraphics[scale=0.5]{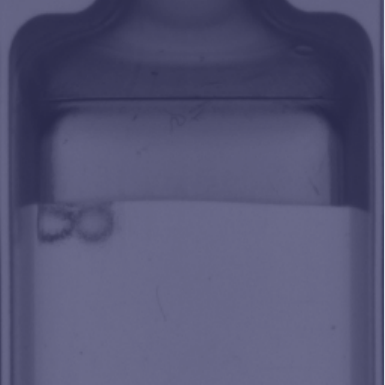}}
			
			\caption{Visual results on real case experiment. First 3 images represent a defect, the last a regular product really difficult to spot.}
			\label{fig:experiment_results}
		\end{figure}
		
		\begin{table}[h]
			\begin{center} {\footnotesize
					\begin{tabular}{lccc}
						\hline\\[0.05ex]
						& \multicolumn{3}{c}{Best results on 30 epochs training}  	\\[1ex]
						& \multicolumn{1}{c}{Best AUROC per image}
						& \multicolumn{1}{c}{Best AUROC per pixel}
						& \multicolumn{1}{c}{Best accuracy}			\\
						\hline\\[0.5ex]
						{\textbf{Vials on meniscus area}} & 0.981 & 0.996 & 0.932 \\[2ex]
						\hline
				\end{tabular} }
			\end{center}
			\caption{\footnotesize Real-case experiment statistics after 30 epochs of training.}
			\label{table:experiment_res}
		\end{table}
		
		With our network we managed to localize those anomalies, with good result, acceptable compared to human and classical algorithms scores.
		These results could be seen in the Figure \ref{fig:experiment_results} and in the table \ref{table:experiment_res}.
		
		\subsection{Ablation study}
		The GRD-Net architecture is analyzed, evaluating the network \textit{generative} model and the loss of the \textit{discriminative} part.
		\subsubsection{\textit{Generative} Model}
		The \textit{generative} sub-net, namely the reconstructive part, was challenged starting from the SoA described in DR{\AE}M paper \cite{zavrtanik2021draem}, in \textit{4.2. Ablation Study - Architecture} sub-section. Adding GAN structure with a residual autoencoder.
		The latter has been tested using a \textit{full-convolutional bottleneck}, with a latent size of $z = 32 \times 8 \times 8$ and a \textit{dense bottleneck}, with a latent size of $z = 2048$.
		As previously shown, best performance were obtained using our GAN architecture, with a fully-convolutional residual autoencoder (CRAE).
		Dense-bottleneck residual autoencoder (DRAE), in the other hand, it is a good alternative, and in some cases is better in anomaly removal task, but is less capable of learning the aleatory areas.
		A good example, shown in Figure \ref{fig:pills_ex}, is pill dataset, whose pills, used as examples, have a random-like dotted reddish texture, that is better reproduced with a fully-convolutional bottleneck.
		
		\begin{figure}
			\centering
			%\hspace{-1.5cm}
			\subfloat[Original pill image ($X$)]{\includegraphics[scale=0.4]{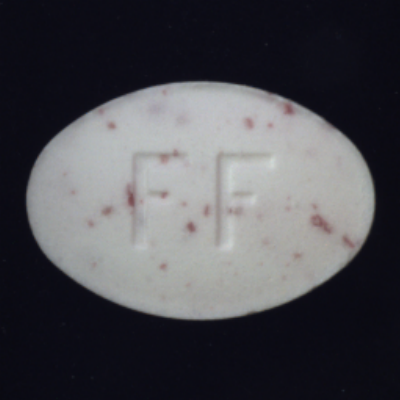}}
			\hspace{0.25cm}\subfloat[Pill image with Perlin noise $X_{(n)}$ ($\hat{X}$)]{\includegraphics[scale=0.4]{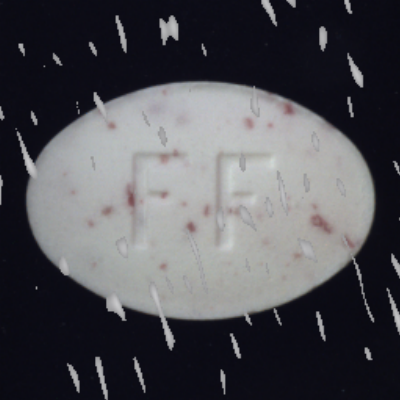}}
			\hspace{0.25cm}\subfloat[Pill image rebuilt by GRD-Net with CRAE ($\hat{X}$)]{\includegraphics[scale=0.4]{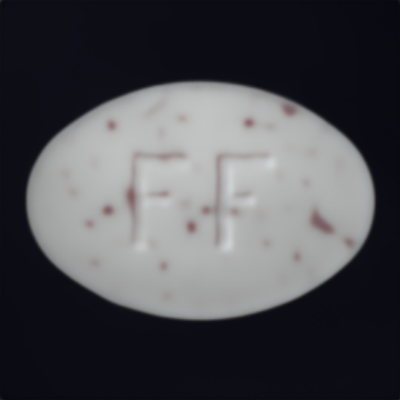}}
			\hspace{0.25cm}\subfloat[Pill image rebuilt by GRD-Net with DRAE ($\hat{X}$)]{\includegraphics[scale=0.4]{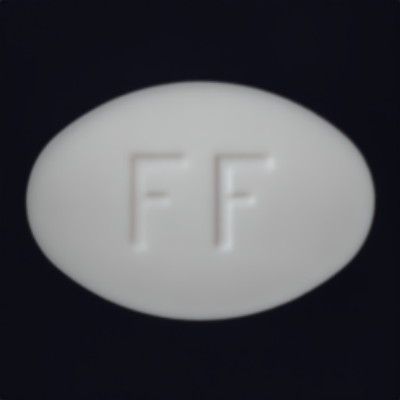}}
			\caption{Pill in picture (a) is the original one from train set, pill in (b) is the original with Perlin noise superimposed. Last two images are the output from the generative subnetwork. (c) is from the full-convolutional residual autoencoder (CRAE), (d) is from the dense-bottleneck residual autoencoder (DRAE). It is clear that the capability to rebuild the original image is much higher in fully-convolutional residual autoencoder (CRAE) networks, especially for the little details on the texture.}
			\label{fig:pills_ex}
		\end{figure}
		
		\subsubsection{\textit{Discriminative} Model Loss}
		Discriminative model loss was originally composed by \textit{Focal Loss} added to the \textit{Crossentropy Overlap Distance Loss} \cite{FraBizCasLam2023-APPIN-IJ,FraBizCasLam2022-ITAL_IA-NW}.
		Initial idea was that the second addendum would help to focus attention of the network only into the ROI area.
		So first idea was:
		\begin{equation}
			\mathcal{L}_{overlap}(\mathcal{A}_{discr}, \mathcal{ROI}_{input}) = {w(1 - {{ \left| \mathcal{A}_{discr} \cap \mathcal{ROI}_{input} \right| } \over {\min ({ \left| \mathcal{A}_{discr}, \mathcal{ROI}_{input} \right| }) }})}.
			\label{eq:loverlap}
		\end{equation}
		with $w \in [0, 1]$.
		Where $\mathcal{L}_{overlap}$ is the contribution of the \textit{Crossentropy Overlap Distance Loss} in the discriminative Loss, $w$ is an hyper-parameter, $\mathcal{A}_{discr}$ is the area mask generated by the discriminative network and $\mathcal{ROI}_{input}$ is the reference ROI
		\begin{equation}
			\mathcal{L}_{FL} = \mathcal{FL}(\mathcal{A}_{discr}, \mathcal{M}_{input}) + \mathcal{L}_{overlap}(\mathcal{A}_{discr}, \mathcal{ROI}_{input}).
			\label{eq:floss}
		\end{equation}
		This loss led the \textit{disciminative} network to focusing on the ROI, but also led to highlight all the ROI area on the heatmap generated as \textit{disciminative} net output.
		This because the (\ref{eq:loverlap}) meant that the $\mathcal{A}_{discr}$ region tends to ${{\mathcal{ROI}_{input}} \cdot {w}}$.
		In order to prevent this issue we performed 4 experiments, with 4 different variations of the $\mathcal{L}_{FL}$:
		\begin{enumerate}
			\item For the first experiment we used the (\ref{eq:floss}).
			\item For the second trial we used the vanilla \textit{focal loss}, but with the intersection, as in the equation (\ref{eq:int_eq}), 	$\mathcal{I} = { \left| \mathcal{A}_{discr} \cap \mathcal{ROI}_{input} \right| } = \mathcal{A}_{discr} \times \mathcal{ROI}_{input}$, as focal loss function input.
			\item For the third experiment we added to the vanilla loss with the input explained in the previous point, the overlap custom loss.
			\item For the fourth, and last, test we negated the overlap function to not intersect $\textbf{1} - \mathcal{ROI}_{input}$
		\end{enumerate}
		Best results, both visually (as shown in Figure \ref{fig:hazelnut_losses}) and numerically (as shown in table \ref{table:ablation}), were obtained using the method 2.
		This due to the tendency to carry $\min ({\mathcal{A}_{discr}} \cap \mathcal{ROI}_{input})$, to be $w$.
		Similar results were obtained on zipper dataset, that was a good benchmark for real-case defects that, on same image, appear both inside and outside the ROI, as illustrated in Figure \ref{fig:results}.
		
		\begin{figure}
			\centering
			%\hspace{-1.5cm}
			
			\raisebox{6\height}{Case 1:}
			\hspace{0.25cm}\subfloat[Original image $X$]{\includegraphics[scale=0.5]{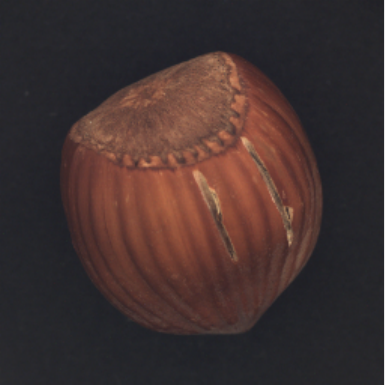}}
			\hspace{0.25cm}\subfloat[Generated heatmap $\hat{M}$]{\includegraphics[scale=0.5]{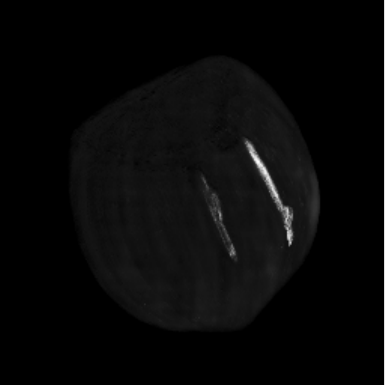}}
			\hspace{0.25cm}\subfloat[Colorized and superimposed $\hat{M}$ on $X$]{\includegraphics[scale=0.5]{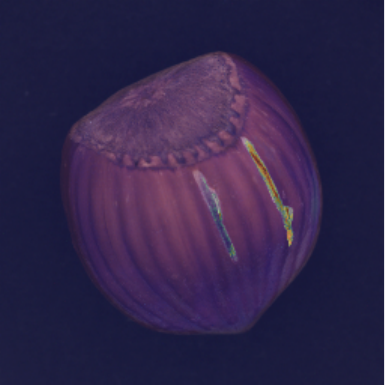}}
			
			\raisebox{6\height}{Case 2:}
			\hspace{0.25cm}\subfloat[Original image $X$]{\includegraphics[scale=0.5]{./img/o_data_1}}
			\hspace{0.25cm}\subfloat[Generated heatmap $\hat{M}$]{\includegraphics[scale=0.5]{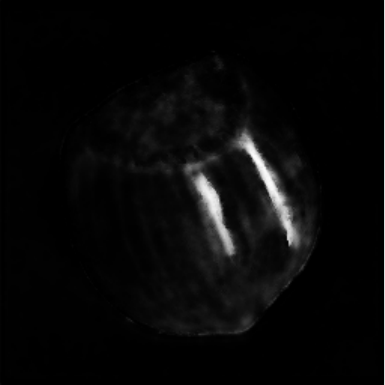}}
			\hspace{0.25cm}\subfloat[Colorized and superimposed $\hat{M}$ on $X$]{\includegraphics[scale=0.5]{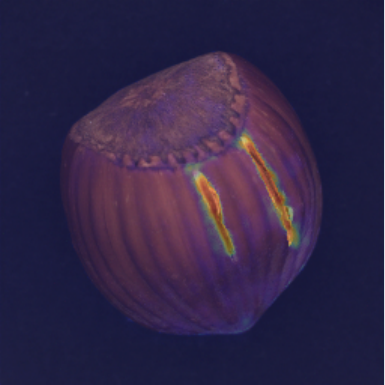}}
			
			\raisebox{6\height}{Case 3:}
			\hspace{0.25cm}\subfloat[Original image $X$]{\includegraphics[scale=0.5]{./img/o_data_1}}
			\hspace{0.25cm}\subfloat[Generated heatmap $\hat{M}$]{\includegraphics[scale=0.5]{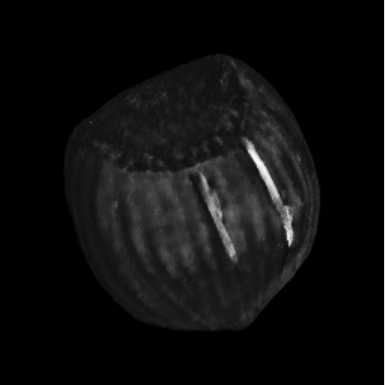}}
			\hspace{0.25cm}\subfloat[Colorized and superimposed $\hat{M}$ on $X$]{\includegraphics[scale=0.5]{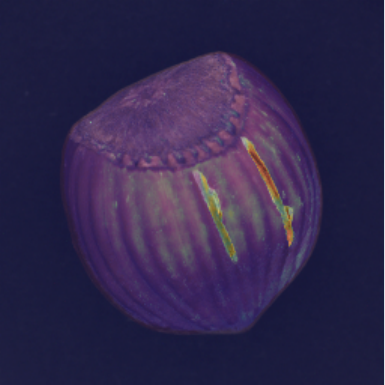}}
			
			\raisebox{6\height}{Case 4:}
			\hspace{0.25cm}\subfloat[Original image $X$]{\includegraphics[scale=0.5]{./img/o_data_1}}
			\hspace{0.25cm}\subfloat[Generated heatmap $\hat{M}$]{\includegraphics[scale=0.5]{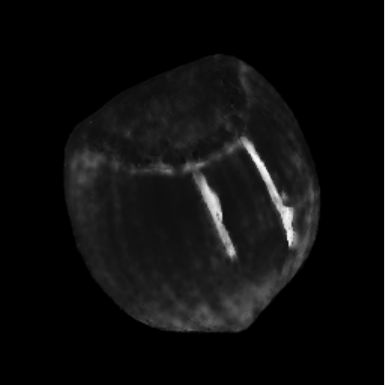}}
			\hspace{0.25cm}\subfloat[Colorized and superimposed $\hat{M}$ on $X$]{\includegraphics[scale=0.5]{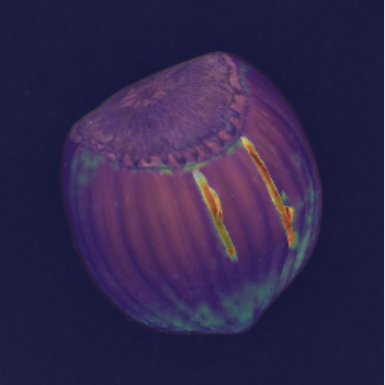}}
			
			\caption{Visual comparison between 4 losses, for the \textit{discriminative} network. Second case is the more clear, that segment better \textit{only} the anomalous areas.}
			\label{fig:hazelnut_losses}
		\end{figure}
		
		\begin{table}[h]
			\begin{center} {\footnotesize
					\begin{tabular}{lccc}
						\hline
						& \multicolumn{3}{c}{AUROC per image (pixel) at 100 epochs}  	\\
						& \multicolumn{1}{c}{AUROC}
						& \multicolumn{1}{c}{AUROX pixel}
						& \multicolumn{1}{c}{Accuracy}			\\
						\hline
						Case 1.
						& 99.4 & 94.3 & 94.6 \\
						Case 2.
						& \textbf{100.0} & \textbf{95.3} & \textbf{100.0} \\
						Case 3.
						& 99.9 & 91.6 & 99.1 \\
						Case 4.
						& \textbf{100.0} & 93.5 & \textbf{100.0} \\
						\hline
				\end{tabular} }
			\end{center}
			\caption{\footnotesize AUROC score after 200 epochs of training per image and pixel.}
			\label{table:ablation}
		\end{table}
	}

\newpage

\section{Conclusions}
	\label{sec: conc_future}
	{
		The aim of this work is to create an anomaly detection network that pay attention mainly to a specific part of an image, to avoid the identification of part of images containing noise defects in the background. This new architecture called Generative-Reconstructive-Discriminative Anomaly Detection with Region of Interest Attention Module network (GRD-Net) is based on two state-of-the-art anomaly detection network: GANomaly and DR\AE M. GDR-Net is composed by a first generative-reconstructive part (GANomaly) trained to identify and reconstruct anomalies, maintaining the non-anomalous regions of the input image. This first sub-model maps the input image to a lower dimension vector using encoder-decoder-encoder sub-networks, which are then used to reconstruct the generated output image. In order to learn joint anomaly inclusion reconstruction and create accurate anomaly segmentation maps, the second network combines the original and reconstructed image. In order to ensure that only the defects present on the surface of the inspected products are considered, in addition to the images of the dataset, the network is also given a segmentation mask that highlights the area of interest (AOI) of the product. This mask is multiplied by the anomaly detection mask generated by the discriminative network to obtain an intersection mask. This contribution is summed to the loss of the network. GRD-Net was tested on all MVTec-AD datasets, on an updated version of the zipper MVTec-AD dataset and on a real industrial dataset provided by company Bonfiglioli Engineering, located in Ferrara (IT).
		Experiments show that GRD-Net performs better than both DR\AE M and GANomaly not only in terms of performance (AUROC) but also in visual terms. In fact, the experiments show that the attention module allows GRD-Net to identify as real defects only those that are in the AOI of the product. In this way, the noise introduced by random variations in the background makes no negative contribution to the performance and reliability of the system created.
	}

\newpage

\bibliographystyle{unsrt}
\bibliography{biblio}

\newpage

\tableofcontents
\printindex
\end{document}